%% file: tacl2021.tex
\pdfoutput=1
\documentclass[11pt,a4paper]{article}
\usepackage{CJKutf8}
\usepackage{booktabs}
\usepackage{multirow}
\usepackage{array,graphicx}
\usepackage{listings}

\usepackage{times,latexsym}
\usepackage{url}
\usepackage[T1]{fontenc}

\usepackage[acceptedWithA]{tacl2021v1}
\usepackage{xspace,mfirstuc,tabulary}

\newif\iftaclinstructions
\taclinstructionsfalse 
\iftaclinstructions
\renewcommand{\confidential}{}
\renewcommand{\anonsubtext}{(No author info supplied here, for consistency with
TACL-submission anonymization requirements)}
\newcommand{\instr}
\fi

\iftaclpubformat 

\else

\fi

\title{Hedges in Bidirectional Translations of Publicity-Oriented Documents}

\author{
  Zhaokun Jiang$^\diamond$
  \and
  Ziyin Zhang$^\dagger$\Thanks{daenerystargaryen@sjtu.edu.cn}
  \\
  \ \\
  $^\diamond$School of Foreign Languages
  \\
  Shanghai Jiao Tong University
  \\
  \\
  $^\dagger$Department of Computer Science and Engineering
  \\
  Shanghai Jiao Tong University
}

\date{}

\begin{document}
\maketitle
\begin{abstract}
  Hedges are widely studied across registers and disciplines, yet research on the translation of hedges in political texts is extremely limited. This contrastive study is dedicated to investigating whether there is a diachronic change in the frequencies of hedging devices in the target texts, to what extent the changing frequencies of translated hedges through years are attributed to the source texts, and what translation strategies are adopted to deal with them. For the purposes of this research, two types of official political texts and their translations from China and the United Nations were collected to form three sub-corpora. Results show that hedges tend to appear more frequently in English political texts, be it original English or translated English. In addition, directionality seems to play an important role in influencing both the frequencies and translation strategies regarding the use of hedges. A noticeable diachronic increase of hedging devices is also observed in our corpus.
\end{abstract}

\input{introduction}
\input{related}
\input{data}
\input{analysis}
\input{discussion}
\input{conclusion}

\bibliography{tacl2021}
\bibliographystyle{acl_natbib}

\end{document}

%% file: introduction.tex
\section{Introduction}

Chinese political discourse has long been an important research topic for interpretation and translation studies. Recent years have witnessed a range of publications from these fields~\citep{2018Gu,2018Liao,2019Fu,2020Li,2022Fu}, with most attention being drawn to a series of official documents and government press conferences, due to their paramount significance in China's diplomatic exchanges with the international society.

Among all the officially-released, public-available written materials, the reports on the work of the government (RWG) are particularly noteworthy, because of their evaluative role on the government's performance in the past year and their guidance on China's future development~\citep{2004Zhang}. The report is delivered by the Premier during the annual National People's Congress (NPC), the highest-level national conference in China, covering topics including the government's works and accomplishments, schedules and targets for the current year, as well as practices of government administration. Generally, it takes the joint efforts of several departments (including the Department of Translation and Interpreting, the Foreign Languages Publishing Administration, and the Xinhua News Agency) led by The Central Compilation and Translation Bureau (CCTB) to translate this report into various languages under unified and strict working processes~\citep{2021Pan}. However, the English version of the RWG may fail to be well-received by the target readership in some cases due to its ``bureaucratic airs'' (``\begin{CJK*}{UTF8}{gkai}官气\end{CJK*}'', ``guan xi'',~\citealp{2004Zhang}), which might be attributed to the differences between the two divergent cultural and linguistic systems~\citep{2022Fu}.

One of the significant differences between the Chinese and English language is the use of hedges. \citet{1973Lakoff} defines hedges as a group of devices to make the propositional content ``fuzzier or less fuzzy'' (195), but further studies indicate that hedges are in fact multifunctional and assume diversified pragmatic and semantic roles, making hedges a widespread linguistic phenomenon. The flexibility and practicality offered by hedges make them an important part of political and diplomatic discourse. Previous studies suggest that compared with English, the Chinese language is relatively ``underhedged'' across genres~\citep{2013Yang,2011Hu,2009Wang,2022Fu}. Given this, hedges can serve as a window to examine and compare the linguistic differences between the Chinese and English political language, as well as the distinction between English translation of Chinese official documents and non-translated English ones.

Also needed to be mentioned is that the Chinese government has shifted the priority of its outward translation from being faithfulness-oriented towards ``target-oriented'' to improve the reception of Chinese diplomatic discourse, and to help the international community better understand the Chinese policies and practices~\citep{2021Pan2,2004Zhang}. Guided by this shifting policy orientation and changing translation norms, it is safe to presume that translators responsible for translating the government's pollical documents to the international audience, though still restricted by a set of fixed working mechanisms, may play a more active role in connecting China with the global community. However, few studies have been done to testify this assumption, and no research has chosen hedges as a particular focus to measure the degree of change. Therefore, the aim of the present study is three-pronged: 1) to examine the diachronic change of hedges being used in the English translations of RWG, and to what extent the differences are attributed to the ST, and 2) to categorize the common translation strategies adopted by translators to deal with the hedging devices.

%% file: related.tex
\section{Literature Review}
\subsection{Hedges and their Pragmatics Uses}
Hedges are commonly understood as a set of lexical or non-lexical devices used to indicate fuzziness, uncertainty and probability~\citep{1973Lakoff,1987Brown,1988Biber}. As hedges are widespread across languages and can serve multiple functions, scholars have conducted various studies to explore their use in both spoken language~\citep{2017Magnifico} and written texts~\citep{2011Abdollahzadeh,2011Hu,2015Yang,2022Yang}, and across disciplines ranging from politics~\citep{2018Ponterotto,2022Fu}, education~\citep{2021Liu}, finance and management~\citep{2010Vazquez,2022Yang}, as well as law~\citep{2017Chaemsaithong}.

The concept of hedges can be traced to \citet[p.195]{1973Lakoff}, who defined the term as ``words whose job is to make things fuzzier or less fuzzy''. He offered a detailed though not exhaustive analysis of their use, and noted that hedges often appeared before a noun phrase or predicate to blur the boundaries between concepts. Although Lakeoff mainly addresses the semantic usage of hedges, later researches tend to approach them from a functional view, establishing connections between hedges and pragmatic functions in different contexts. \citet[p.70]{1987Brown} discussed how hedges can intervene in face-threatening acts to protect the speaker's positive face and negative face. In a similar vein, \citet[p.185]{1990Holmes} regard hedges as a speech strategy in line with the politeness principle. \citep{2005Hyland} treated hedges as being part of a broader and systematic framework. To be specific, he proposed an interpersonal metadiscourse model consisted of two subgroups— ``interactive'' and ``interactional'', and put hedges into the ``interactional'' subgroup, whose purpose is to ``involve the reader in the text'' (49). In his view, hedges demonstrate the language user's cautiousness against absolute proposition, and serve to modify attitudes toward the truth value of propositional message.

However, there has been no consensus in terms of what should be counted as hedges, and how to classify them. \citet{1982Prince} put forward a classification of hedges comprising two major categories (\textit{approximators} and \textit{shields}), treating hedges as both semantic and pragmatic devices. Following \citep{1982Prince}, \citet{1994Salager-Meyer} further subdivided hedges into \textit{shields}, \textit{approximators}, \textit{emotionally-charged intensifiers} (e.g. ``particularly encouraging'', ``unexpectedly'', ``extremely difficult''), and \textit{compound hedges} (e.g. ``it may suggest that'', ``it would seem likely that''). Mainly focusing on grammatical features, \citet{1998Hyland} proposed a model of hedges consisted of four categories: \textit{model verb}; \textit{epistemic lexical verb}; \textit{epistemic adjective, adverb, and noun}; \textit{phraseological expression}. It seems that there is also distinction between hedges often used in spoken language and those seen in written texts. Researchers have identified a set of hedges specific to oral discourse. For example, \citet{1987Schiffrin} regarded the word ``well'' as a discourse marker to introduce new topics, leave time to pause, provide answers to questions, and express disagreement. \citet{1998Jucker} examined how the hedge ``you know'' contributed to establishing connection with the speech receiver and encouraging positive reception of the information offered by the speech giver.

Moreover, some scholars mentioned that in a broader sense, hedging could also be a discourse strategy, not limited to a set of specific words or phrases. \citet{2003Martin} identified four basic hedging strategies, including \textit{Indetermination, Camouflage, Subjectivization and Depersonalization}. Similarly, \citet{2018Ponterotto} viewed hedging as referring to ``virtually any discourse strategy aimed at avoiding explicit position when ascertaining the truth conditions of reported facts or events'' (178).

\subsection{Intercultural Studies of Hedges}
It is noteworthy that research on hedges has attracted an increasing number of scholars from diversified cultural background in recent years, which greatly expands the scope of this field and offers fresh understanding into the various uses of hedges in different socio-cultural contexts. For example, \citet{2015Yang} conducted a case study that taps into a special Chinese hedge ``kongpa''. Based on Mandarin broadcast transcripts, the study observed that although ``kongpa'' roughly equates to ``being afraid'' in English, it was rarely linked with fear or cautiousness in spoken Chinese. Rather, the word was often used neutrally or even positively for interpersonal purposes, to protect face of the speaker or the hearer, to avoid self-praising, or to criticize others in a euphemistic manner.

Also noticeable is that much attention has been drawn to comparative studies of hedges from a cross-cultural perspective. For instance, \citet{2011Hu} adopted a corpus-based approach to instigate the similarities and distinctions of hedging patterns between English and Chinese academic abstracts. Results of their empirical study showed that the frequency of hedges in English-medium abstracts was significantly more than that in Chinese-medium abstracts, and the authors attributed this difference to divergent rhetorical conventions of the two languages, as well as different conceptions of scientific publication. \citet{2011Abdollahzadeh} examined the use of hedges in academic papers written in English by Anglophone speakers and Iranian scholars. Based on 60 conclusion parts extracted from two groups of articles, the author found that hedges appear more frequently in papers written by Anglophone speakers. \citet{2019Shafqat} focused on journalistic English, conducting a comparative study of hedges in a Pskistani English newspaper and an European English newspaper according to the classification of \citet{2005Hyland}. They used statistical methods to test their assumptions about the use of hedges in the two groups of newspaper, which are supported by their empirical results: hedges of all types appear significantly more frequent in the European English newspaper than in the Pakistani one.

These intercultural studies offer great insight into the differentiated use of hedges in multiple cultures and in various disciplines. However, the bulk of literature has been centered on a few dominant languages, especially English and Spanish, leaving ample space to conduct research on less dominant languages. Also, as comparative studies in this field have gained momentum, results from many studies show that hedges tend to be used more frequently in English than in other languages~\citep{2013Yang}. Nevertheless, few studies provide detailed explanation regarding why this phenomenon exists. As language is not merely a linguistic product, more studies are awaited to dissect the different use of hedges from a socio-cultural perspective.

\subsection{Translation of Hedges in Political Settings}
Political discourse, whether in spoken or written form, is generally planned, formal, well-prepared, and strategically-expressed in advance~\citep{2019Gribanova}, because they are open to close scrutiny by both domestic and international audience. Because of this, proper deployment of hedges is often a necessity in political discourse to both satisfy the expectation of the target audience and protect the image and interest of the massage senders.

Although hedges seem to have a natural connection with politics, research on the use of hedging devices in political texts is relatively limited. Among the few related studies, president's speeches have attracted most attention. For example, \citet{2010Fraser} was particularly interested in how politicians resorted to hedging when facing challenging questions from journalists during press conference. He chose the former US president Bush's 2007 Press Conference as his source of research material. Another study conducted by \citet{2018Ponterotto} analyzed hedging strategies adopted by another former president Obama in his political interviews. The author suggested that Obama used various hedging moves and strategies to evade explicit answers on sensitive issues, and his artful use of hedges contributed to a cautious speech style.

Translational research focusing on hedges is even fewer. Because of the small number of studies in this specific field, the following review is not strictly confined to translation studies, but includes interpreting studies as well. In \citet{2017Magnifico}, the authors drew on speeches and their corresponding interpretation transcripts at the European Parliament to examine whether gender differences existed in the linguistic output during spontaneous interpretations. The authors found that female interpreters used significantly more hedging devices than their male counterparts. \citet{2022Fu} compared the frequency of hedges in interpreted Chinese political press briefings with that in spontaneous English speeches, and found that interpreted speeches contained significantly fewer hedges, and that the range of hedging devices were less diversified than speeches made by Anglophone speakers in similar political settings. \citet{1987Schiffrin} investigated how hedges were translated from English to German in Tony Blair's speech during the 1995 Labour Party Conference. The author identified three translation strategies to render hedges in a different cultural context, and found that hedging devices could modify the precision of discourse, facilitate interpersonal exchanges with other politicians, show politeness, and build a positive self-image, fulfilling both semantic and pragmatic functions. In a newly published study, \citet{2022Aljawadi} examined the English-Arabic translation of hedges in the former US president Donald Trump's public speeches and interviews related to the Covid-19 pandemic. Based on Fraser's classifications, the study found that hedges were most frequently used in the context of negation.

The previous literature of hedges from a translational perspective sheds light on the contributions of hedging devices in the cross-cultural delivery of political texts. Nevertheless, all of these studies only addressed single-direction translation of hedges (e.g. Chinese-to-English, English-to-German). No study has been done to deal with their bi-directional translation in similar political settings. In other words, there has not been a complete picture of how hedges are used to communicate with the domestic and international audience, and whether differences exist between the SL-TL and TL-SL translation of hedging devices, leaving a gap to be filled.

%% file: data.tex
\section{Data and Methodology}
\begin{table*}[]
    \centering
    \scalebox{0.9}{
    \begin{tabular}{cccm{5cm}<{\centering}c}
    \toprule
         & \multicolumn{2}{c}{ST} & \multicolumn{2}{c}{TT} \\
         & Items & Tokens & Items & Tokens \\
    \midrule
        \multirow{6}{*}{RWG1} & Report in 2000 & 7657 & Translated report in 2000 & 13079 \\
         & Report in 2001 & 7605 & Translated report in 2001 & 13210 \\
         & Report in 2002 & 7454 & Translated report in 2002 & 12865 \\
         & Report in 2003 & 11371 & Translated report in 2003 & 18921 \\
         & Report in 2004 & 8242 & Translated report in 2004 & 13877 \\
         & Total & 42329 & & 71952 \\
    \midrule
        \multirow{6}{*}{RWG2} & Report in 2018 & 14798 & Translated report in 2018 & 24201 \\
         & Report in 2019 & 11707 & Translated report in 2019 & 19877 \\
         & Report in 2020 & 8291 & Translated report in 2020 & 16059 \\
         & Report in 2021 & 10661 & Translated report in 2021 & 19147 \\
         & Report in 2022 & 8895 & Translated report in 2022 & 16491 \\
         & Total & 54352 & & 95775 \\
    \midrule
        \multirow{6}{*}{UNAR} & Report in 2018 & 9650 & Translated report in 2018 & 13620 \\
         & Report in 2019 & 9687 & Translated report in 2019 & 16406 \\
         & Report in 2020 & 8962 & Translated report in 2020 & 15175 \\
         & Report in 2021 & 7640 & Translated report in 2021 & 12194 \\
         & Report in 2022 & 8432 & Translated report in 2022 & 13869 \\
         & Total & 44371 & & 71264 \\
    \bottomrule
    \end{tabular}}
    \caption{Basic information of the three sub-corpora.}
    \label{tab:1}
\end{table*}

\subsection{Corpus Data}
This study adopts a corpus-based approach, comprising three parallel sub-corpora: 

(1) the original reports on the work of the government and their English translations from 2000 to 2004 (RWG1); 

(2) the original reports on the work of the government and their English translations from 2018 to 2022 (RWG2); 

(3) the English United Nations annual reports and their Chinese translations from 2018 to 2022 (UNAR). All the textual data from the first two sub-corpora are directly downloaded from the official website of China's state council, which provides documents both in Chinese\footnote{\url{http://www.gov.cn/premier/lkq_jh.htm}}, and in English\footnote{\url{http://english.www.gov.cn/premier/speeches/}}. Data in the third sub-corpus come from the United Nations' official website\footnote{\url{https://www.un.org/annualreport/zh/index.html}}, on which annual reports for the latest five years are available in English, Chinese and other official languages.

The UN annual reports, delivered by the general secretary during the general assembly, are highly similar to the RWGs in both register and topics, and thus are used as the comparable corpus in this study. Reports on the work of the government are the most comprehensive record of works done during the past year, cover a wide range of topics of national importance, including economic development, social welfares, ecological civilization, political reforms, diplomatic affairs, and cultural undertakings. Similarly, the UN annual reports are usually delivered by the general secretary during the general assembly, recording all the major achievements accomplished in the established fields of priority, including promotion of sustainable economic growth, peace keeping and security, human rights, as well as law and justice.

Since the downloaded UN annual reports are pdf files, they are first transformed into machine- readable txt files. Following that, EmEditor is used to re-arrange the format, eliminate extra spaces between words, delete blank lines, and replace wrong punctuations. Next, ABBYY Aligner 2.0 is used to align all the original texts and their translations. After the automatic alignment by the software, manual adjustment is done to ensure that the source texts and translated texts are accurately matched.

\begin{CJK*}{UTF8}{gkai}
\begin{table*}[]
    \centering
    \scalebox{0.9}{
    \begin{tabular}{ccm{11.5cm}}
    \toprule
        \multirow{7}{*}{\textbf{\textit{approximators}}} & \multirow{4}{*}{\textbf{\textit{adaptors}}} & some, several, a portion of, certain, somewhat, quite, kind of, sort of, almost, can, could, might, may \\
        \cmidrule{3-3}
         & & 一些 (yi xie)，有些 (you xie)，部分 (bu fen)，有的 (you de)，有所 (you suo)，几 (ji)，有点 (you dian)，几乎 (ji hu)，可能 (ke neng)\\
         \cmidrule(l{0.5em}){2-3}
         & \multirow{3}{*}{\textbf{\textit{rounders}}} & approximately, about, roughly, around，nearly\\
         \cmidrule{3-3}
         & & 大约 (da yue)，大概 (da gai)，围绕 (wei rao)，上下 (shang xia)，左右 (zuo you)，近 (jin), 不到 (bu dao)\\
    \midrule
        \multirow{11}{*}{\textbf{\textit{shields}}} & \multirow{5}{2cm}{\textbf{\textit{plausibility shields}}} & I/we think, I/we (firmly) believe, I/we suppose, I/we wonder, from my/our understanding, to my/our knowledge\\
        \cmidrule{3-3}
        & & 我/我们相信 (wo/wo men xiang xin)，我/我们觉得 (wo/wo men jue de)，我/我们认为 (wo/wo men ren wei)，我/我们想 (wo/wo men xiang)，据我/我们所知 (ju wo/wo men suo zhi)\\
        \cmidrule(l{0.5em}){2-3}
        & \multirow{6}{2cm}{\textbf{\textit{attribution shields}}} & it seems, it appears, according to, accordingly, in accordance with, it is said, it is reported, presumably, in line with, as the (old/Chinese) saying goes, -based, based on\\
        \cmidrule{3-3}
        && 似乎 (si hu)，好像 (hao xiang)，(根)据 ((gen) ju)，依(照) (yi (zhao)), 据说 (ju shuo)，据报道 (ju bao dao)，表明 (biao ming)，基于 (ji yu), 考虑(到) (kao lv (dao))\\
    \bottomrule
    \end{tabular}}
    \caption{Taxonomy of hedges adapted and expanded from \citet{1982Prince}.}
    \label{tab:2}
\end{table*}
\end{CJK*}

\subsection{Research Methods}
The taxonomy of hedges in this study is adapted from \citet{1982Prince}, which consists of two major categories and four sub-categories, treating hedges as a set of expressions with both semantic and pragmatic functions. In their model, \textit{approximators} can be divided into \textit{adaptors}, which are mainly used to downgrade the absoluteness of propositional information (e.g. somewhat), and \textit{rounders}, which indicate a range or degree around which the proposition can be valid (e.g. about, may). \textit{Shields} also contain two sub-categories: \textit{plausibility shields} are used to confirm that the proposition is made by the speaker (e.g. I believe), and \textit{attribution shields} soften the truth value of a proposition by indicating that the utterance is quoted from others (e.g. ``according to'').

However, since the original taxonomy proposed in \citet{1982Prince} is mainly targeted at the spoken language, it may not fit the purpose of the current study. For this consideration, we drew from both their framework and other previous research to find as many hedges as possible and put them under appropriate category. In addition, as there is no systematic classification of hedges in the Chinese language, we have to first search with all the English hedges and find their Chinese counterparts with the help of our parallel corpora. In this manner, we can find more hedging devices not included in our original taxonomy for both the language pairs.

After that, concordance lines containing these hedges will be extracted, and their corresponding expressions in the ST or TT can also be identified via the parallel sub-corpora. Since the extraction process is based on searching for and matching hedges according to the taxonomy without considering the context, some extracted ``hedges'' are not actually used for the purpose of hedging. For example, in ``Development system reform is about becoming much more effective, well- coordinated, transparent and accountable to better assist countries in implementing the 2030 Agenda for Sustainable Development'', ``about'' is not used as a hedge though it is extracted in the first place. These ``false hedges'' will not be considered in our further analysis. Next, the raw and normalized frequency of hedges in the three sub-corpora will be counted, and chi-square testing will be conducted to investigate the differences in the deployment of hedges in both C-E and E-C translations of political texts, and the diachronic patterns of hedges in C-E reports on the work of the government. Following that, translation strategies will be categorized, and textual examples will be discussed in details to analyze how hedging devices in the ST are expressed in TT.

%% file: analysis.tex
\section{Data Analysis}
\begin{table*}[]
    \centering
    \begin{tabular}{ccccccc}
    \toprule
        \multirow{2}{*}{Hedges} & \multicolumn{2}{c}{RWG1} & \multicolumn{2}{c}{RWG2} & \multicolumn{2}{c}{UNAR}\\
        \cmidrule{2-7}
        & ST (zh) & TT (en) & ST (zh) & TT (en) & ST (en) & TT (zh)\\
    \midrule
        Raw freq. & 217 & 249 & 373 & 437 & 331 & 324 \\
        Norm freq. (per 1000) & 5.13 & 3.47 & 6.86 & 4.58 & 4.66 & 4.55\\
    \bottomrule
    \end{tabular}
    \caption{Raw and normalized frequencies of hedges in the three sub-corpora.}
    \label{tab:3}
\end{table*}

In this section, we present major results of our data analysis showing the differences in the use of hedges both between the language pairs synchronically and within each language across time. To be specific, raw and normalized frequencies of hedges in the ST and TT in RWG1, RWG2, and UNAR will be counted respectively and then compared to see if the translation process brings about significant difference in the frequencies of hedging devices in the TT, and whether directionality influences the occurrences of hedges. Next, English hedges in the TT of RWG1 will be counted and compared with those in RWG2 to examine whether a diachronic change regarding the use of hedging expressions exists between the two sub-corpora. Finally, translation strategies will be identified to reveal how institutional translators deal with the hedging devices.

\subsection{Occurrence of Hedges in the Original and Translated Political Texts}
Raw and normalized frequencies (per 1000 tokens) of hedges in the three sub-corpora are presented in Table 3. As we can see, in both RWG1 and RWG2, hedges occur more frequently in the TT than in the ST, which indicates that a certain portion of hedges not seen in the STs are added in the process of translation. In contrast, a reverse pattern is observed in UNAR, in which hedging devices appear more frequently in the ST. These results suggest that hedges tend to appear more frequently in English than in Chinese, which coincide with the previous studies~\citep{2010Vazquez,2011Hu,2021Gong,2013Yang}.

Another interesting finding is that original English political texts tend to contain more hedges than the translated English texts, at least in the case of our three sub-corpora. To illustrate, the ST in UNAR contains significantly more hedges than translated English texts in RWG2 ($\chi^2$=88.48, $df$=1, $p$<0.001), and the difference also amounts to significance between the ST in UNAR and the TT in RWG1($\chi^2$=46.70, $df$=1, $p$<0.001). The data indicate that there might be a systematic difference between the original English and translated English concerning the use of hedges.

Also worth mentioning is that directionality is an important factor for the differences between the STs and TTs regarding the frequencies of hedges. More hedges are seen in C-E translation compared with E-C translation. In RWG1 and RWG2, the ST-TT difference in the number of hedges is 17.2\% and 14.7\% respectively, while the percentage for UNAR is merely 2.1\%. In other words, the Chinese translations of UNAR are more closely aligned with the STs in terms of hedges, while the English translation of hedges in RWG1 and RWG2 are less faithful to the STs .

\begin{table*}[]
    \centering
    \begin{tabular}{ccccc}
    \toprule
        \multirow{2}{*}{Categories} & \multicolumn{2}{c}{RWG1 (TT)} & \multicolumn{2}{c}{RWG2 (TT)} \\
        & Raw freq. & Norm freq. & Raw freq. & Norm freq. \\
    \midrule
        Adaptors & 103 & 1.44 & 91 & 0.95 \\
        Rounders & 39 & 0.54 & 61 & 0.64 \\
        Plausibility Shields & 2 & 0.03 & 0 & 0 \\
        Attribution Shields & 105 & 1.46 & 285 & 2.99\\
        Total & 249 & 3.47 & 437 & 4.58\\
    \bottomrule
    \end{tabular}
    \caption{Raw and normalized frequencies (per 1000 words) of hedges in RWG1 and RWG2.}
    \label{tab:4}
\end{table*}

\subsection{Diachronic Change Regarding the Use of Hedges in the TTs}
To examine whether there are noticeable changes concerning the deployment of hedges in the translations of Chinese political documents through years, we have counted the raw and normalized frequencies of hedging devices according to our taxonomy in the target text in RWG1 and RWG2. On the whole, there are significantly more hedges in RWG2 than in RWG1 ($\chi^2$=12.25, $df$=1, $p$<0.001), which attests the observation that the translation of Chinese political documents has become more target-oriented to improve their reception in the international community~\citep{2021Pan2,2004Zhang}. The increased number is partly attributed to the increase of hedges in the ST, or what is called ``source text stimuli'' by \citet{2017Magnifico}. This is easily understandable, since ``faithfulness'' has been the constant priority for Chinese institutional translators working for the government~\citep{2021Pan2}. Still, a noticeable portion of the increased hedges are not translations from the corresponding hedges in the ST, and have to be explained by factors other than the ST. In other words, these ``extra'' hedges are added by the translators out of various purposes, and thus can be seen as the result of the translators' subjectivity.

As revealed in Table 4, attribution shields are the most dominant hedges across the four categories, followed by adaptors and rounders, while plausibility shields seldom appear in both the two sub-corpora. It is also noted that differences exist in the preferred categories of hedges between RWG1 and RWG2. Specifically, RWG1 displays significantly more adaptors than RWG2 ($\chi^2$=33.00, $df$=1, $p$<0.001), while RWG2 contains nearly three times as many attribution shields as RWG1 ($\chi^2$=34.35, $df$=1, $p$<0.001). The differences in other two categories do not amount to significance. Particularly noteworthy is that plausibility shields are rarely seen in both the two sub- corpora. This might be due to their inconsistency with the stylistic features of the Chinese political texts, which are usually formal, firm, and in lack of subjectivity~\citep{2008Wang}.

\subsubsection{Adaptors}
Table 5 displays the frequencies of adaptors in the TT of RWG1 and RWG2. ``Some'' and ``can'' are the most dominant devices in both the two sub-corpora, which constitute more than two-thirds of all the adaptors. Our results are not aligned with \citet{1982Prince}, which identified ``sort of'' and ``kind of'' as the most frequently used adaptors. Such disparity is caused by the difference in the typology of texts: while ``sort of'' and ``kind of'' are widespread in the spoken discourse~\citep{1988Biber,1973Lakoff,1982Prince}, they are less likely to be seen in the official political documents.

\begin{table}[]
    \centering
    \begin{tabular}{ccc}
    \toprule
        \textbf{Adaptors} & RWG1 (TT) & RWG2 (TT)\\
    \midrule
        some & 56 & 37 \\
        can & 27 & 22 \\
        certain & 10 & 11\\
        may & 7 & 8\\
        almost & 3 & 6\\
    \bottomrule
    \end{tabular}
    \caption{High-frequency adaptors.}
    \label{tab:5}
\end{table}

\begin{quotation}
\noindent Example 1 (RWG1 03/16/2018)

\textbf{\textit{Some}} enterprises, particularly small and medium ones, are finding it tough going. Growth in private investment is weak; \textbf{\textit{some}} regions still face considerable downward economic pressure, and risks and potential dangers in the financial and other sectors are not to be ignored. Poverty alleviation remains a formidable task; agriculture is not based on a strong foundation. The disparities in development between rural and urban areas, between regions, and in income distribution remain substantial. Serious and major workplace accidents happen all too often. People still have a lot of complaints about air quality, environmental sanitation, food and drug safety, housing, education, healthcare, employment, and elderly care. The transformation of government functions has not yet reached where it should be. In government work there are places where we fall short. \textbf{\textit{Some}} reform measures and policies have not been fully implemented. \textbf{\textit{Some}} officials are weak on awareness that they are there to serve and must uphold the rule of law, and \textbf{\textit{some}} lack commitment to their work and willingness to bear the weight of responsibility. Bureaucratism and the practice of formalities for formalities' sake exist to varying degrees. There are many complaints from the people and the business sector about the difficulty of accessing government services and the excessive array of charges. In \textbf{\textit{some}} sectors misconduct and corruption are still a common problem.
\end{quotation}

The frequent occurrence of ``some'' in the TTs is fully reflected in this extract, which contains 6 ``some'' out of 220 words. Most of them follow the ``some + countable nouns/ uncountable noun'' structure, indicating that ``some'' is used to restrict the scope of the mentioned object, making the propositions more generalized.

\subsubsection{Attribution Shields}
\begin{table}[]
    \centering
    \scalebox{0.9}{
    \begin{tabular}{m{3cm}cc}
    \toprule
        \textbf{Attribution Shields} & RWG1 (TT) & RWG2 (TT) \\
    \midrule
        -based & 0 & 95\\
        based on & 10 & 64\\
        in accordance with & 69 & 78 \\
        according to & 25 & 47\\
        accordingly & 1 & 0\\
        as the Chinese saying goes & 0 & 1\\
        Total &105&285\\
    \bottomrule
    \end{tabular}}
    \caption{Frequency distribution of attribution shields.}
    \label{tab:6}
\end{table}

Table 6 lists the raw frequencies of attribution shields in the two sub-corpora. It is noteworthy that the distribution is heavily skewed, with only a small portion of attribution shields being frequently deployed, while the others only appear occasionally or not appear at all. Our results give support to ``simplification'' as a translation universal in the target text~\citep{2019Baker,2004Chesterman}, which refers to a general pattern that information in the ST is often simplified during the process of translation. The phenomenon of simplification can be observed from the lexical, syntactic, grammatical, as well as other aspects. On the lexical level, the translated text generally features less lexical variety and more occurrence of high-frequency items~\citep{2019Baker}.

It is also noticeable that although most attribution shields tend to occur at similar frequencies, ``-based'' and ``based on'' appear significantly more times in RWG2 than in RWG1 ($\chi^2$=46.27, $df$=1, $p$<0.001; $\chi^2$=8.35, $df$=1, $p$=0.004). While the ``NOUN-based'' pattern does not appear at all in RWG1, and ``based on'' only appears 10 times, they occur 95 and 64 times respectively in RWG2, representing a marked increase.

\begin{table*}[]
    \centering
    \scalebox{0.95}{
    \begin{tabular}{rcl}
    \toprule
        respects, deepen reform, advance & \textbf{law-based} & governance, and strengthen Party \\
        have reformed and improved the & \textbf{market-based}& exchange rate mechanism and kept\\
        qualification examinations for some & \textbf{license-based}&professions. We will support the\\
        be better integrated and& \textbf{performance-based}&management will be strengthened\\
        the internet and other & \textbf{IT-based}&approaches to strengthen oversight\\
        including developing at-home, & \textbf{community-based}&and mutual-aid elderly care\\
        enforced in a strict,& \textbf{procedure-based}&impartial, and civil manner\\
        We will strengthen& \textbf{auditing-based}&oversight. We will consolidate and\\
        full support in exercising& \textbf{law-based}&governance and in their efforts\\
        We will promote credit& \textbf{rating-based}&regulation and the Internet Plus\\
    \midrule
        a new model of social governance & \textbf{based on}&collaboration, co-governance, and\\
        work on building a government & \textbf{based on}&the rule of law, and\\
        and create a business environment & \textbf{based on}&rule of law that the\\
        continue to promote opening up & \textbf{based on}&flows of goods and\\
        greater emphasis to opening up & \textbf{based on}&rules and related institutions\\
        core and an international order& \textbf{based on}&international law. China is\\
        while also pursuing progress. & \textbf{Based on}&China’s realities, we refrained\\
        adjustments and improvements & \textbf{based on}&new developments to reinforce\\
        effective tax-and-fee reduction steps& \textbf{based on}&local conditions and in the\\
    \bottomrule
    \end{tabular}}
    \caption{Concordance lines of ``NOUN-based'' and ``based on'' in RWG2 (TT).}
    \label{tab:9}
\end{table*}

Based on these observations, we hypothesize that the soaring occurrence of the two hedges during the past two decades are at least partly attributed to the language contact with English. To test our hypothesis, we searched the two expressions on BNC and COCA, and found that they appear at a much higher frequency in these two corpora compared with RWG1 ($\chi^2$=977.20, $df$=1, $p$<0.001), but appear significantly less frequently compared with RWG2 ($\chi^2$=11358.5, $df$=1, $p$<0.001). In other words, although these hedging devices are rarely used two decades ago, they have managed to be included in the lexicon of publicity-oriented English translated from Chinese, and gained increasing significance until their frequencies surpass those of the original English to a large extent.

\begin{table}[]
    \centering
    \begin{tabular}{ccc}
    \toprule
        Corpora & BNC & COCA \\
    \midrule
        NOUN-based & 8437 (0.09) & 84381 (0.08) \\
        based on & 11337 (0.12) & 141241 (0.14) \\
    \bottomrule
    \end{tabular}
    \caption{Frequency (per 1000 words) of ``NOUN-based'' and ``based on'' in BNC and COCA.}
    \label{tab:7}
\end{table}

In terms of actual usage, it is easily noticeable that nouns in ``NOUN-based'' expressions tend to be well-known places in both BNC and COCA, such as ``California'', ``London'', ``Massachusetts'', ``New York'', and ``Washington''. They can also be ``computer'' or things related to computers (such as the computer system ``Unix'', and ``web'').

\begin{table}[]
    \centering
    \begin{tabular}{ccc}
    \toprule
        Corpora & Patterns & Frequency \\
    \midrule
        \multirow{10}{*}{BNC} & California-based & 355\\
        & London-based & 280 \\
        & computer-based & 184\\
        & Unix-based & 143\\
        & Massachusetts-based & 134\\
        & community-based & 113 \\
        & RISC-based & 113\\
        & UK-based & 92\\
        & land-based & 91 \\
        & broad-based & 90\\
    \midrule
        \multirow{10}{*}{COCA} & evidence-based & 3185\\
        & community-based & 2669\\
        & web-based & 2241 \\
        & school-based & 2031\\
        & faith-based&1852\\
        & New York-based & 1547\\
        & broad-based & 1247\\
        & Washington-based & 1047\\
        & Atlanta-based & 1016\\
        & Chicago-based & 988\\
    \bottomrule
    \end{tabular}
    \caption{``NOUN-based'' patterns in BNC and COCA.}
    \label{tab:8}
\end{table}

\begin{table*}[]
    \centering
    \begin{tabular}{cccccc}
    \toprule
        Sub-corpora & Retention & Addition & Omission & Modification & Total\\
    \midrule
        RWG1 & 215 & 18 & 3 & 19 & 252 \\
        RWG2 & 368 & 42 & 5 & 32 & 442 \\
        UNAR & 324 & 0 & 7 & 0 & 331\\
    \bottomrule
    \end{tabular}
    \caption{Handlings of hedges in the three sub-corpora.}
    \label{tab:10}
\end{table*}

Interestingly, a close examination of the concordance lines in our RWG2 reveals a different picture. As displayed in Table 9, ``-based'' and ``based on'' tend to co-occur with ``law'', ``market'', ``license'', ``performance'', ``procedure'', ``condition'', ``need'', ``reality'', and so on. The frequent cooccurrence of the two attribution shields with these words can lend support to the propositional contents in officially released government documents, making them more credible, objective, and authoritative. The different usage of ``NOUN-based'' and ``based on'' in the original English and translated English in our corpus indicates that although these two expressions are borrowed from the English world, adaption and variation happen to their actual usage as they enter the C-E translational system in the politics-specific genre.

\subsection{Handling of Hedging Devices in Translation}
This section aims to gain deeper insight of how translators deal with hedges in the STs and the possible factors behind different handlings through both quantitative analysis and qualitative analysis of several examples between the language pairs. Careful examination of all the aligned concordance lines containing hedges suggests that the majority of hedges are retained and translated faithfully in the TTs. For the cases where hedges are not retained, addition and modification are most frequently adopted. In contrast, the strategy of omission is only used occasionally.

Also noteworthy is the relationship between directionality and the proportion of retention. To be specific, for the C-E translation in RWG1 and RWG2, the retention rate of hedges is 83.26\% and 85.32\% respectively, while in UNAR, the number is up to 97.86\%. This indicates that the use of hedges is more strictly in line with the STs for E-C translation, but less strictly for C-E translation, at least in the case of our data.

\subsubsection{Addition}
The first identified method to deal with hedges is addition, which refers to that the translator intentionally adds hedging devices in the TT when there are no traces of corresponding hedges in the ST. For the cases where hedges are not retained, addition serves as the dominant method, taking up 9.5\% in RWG1 and 7.1\% in RWG2. This strategy is generally adopted to restrict the scope of mentioned objects (see example 1) or to lower the absoluteness of the proposition.

\begin{CJK*}{UTF8}{gkai}
\begin{quotation}
\noindent Example 1 (2003 RWG)

ST: 坚持不懈地开展反腐败斗争，大力纠正部门和行业不正之风，依法惩处了一批违法违纪 的腐败分子。

Gloss: We made unremitting efforts to combat corruption, rectify unhealthy tendencies in the departments and trades and punish according to law quite a few corrupt elements.

TT: We made unremitting efforts to combat corruption, rectify unhealthy tendencies in some departments and trades and punish according to law quite a few corrupt elements.
\end{quotation}
\end{CJK*}

Example 1 demonstrates the resolution of the Chinese government to combat corruption and improve discipline in all the government departments and enterprises amid the rampant malpractices among some senior officials and managers in some state-owned enterprises. In fact, several measures are reported in details about how the government aims to cope with corruption in the previous parts of the report. However, widespread corruption is definitely detrimental to China's image in the international society, thus the translator adds the hedge ``some'' before ``departments and trades'' in the TT, to the effect of limiting the corruption issue to only a portion of all the departments and trades.

\subsubsection{Omission}
Omission is the least adopted method in RWG1 and RWG2. This might be attributed to the systematic difference between Chinese and English regarding the use of hedges: in general, Chinese tends to be ``under-hedged'' across all genres than English, which is supported by many previous studies~\citep{2010Vazquez,2011Hu,2021Gong,2013Yang}. Therefore, when the political documents originally written in Chinese are to be translated into English, hedges are often added rather than omitted. In contrast, for E-C translation, omission is preferred, which may explain why omission is the only adopted translation method other than retention in the sub-corpus UNAR. As suggested by example 2, the omission of hedges generally serves to strengthen the propositional content.

\begin{CJK*}{UTF8}{gkai}
\begin{quotation}
\noindent Example 2 (2018 UN annual report)

ST: The incidence of homicides and violence relating to organized crime remains high in many regions in the world and, when linked to the illicit trafficking of arms and commodities, can derail efforts towards peace, human rights protection and sustainable development.

TT: 在世界许多地区，与有组织犯罪有关的凶杀和暴力行为发生率仍然很高，如果这些行为 与非法贩运武器和初级商品行为牵扯在一起，会破坏实现和平、保护人权和可持续发展的努力。

BT: The incidence of homicides and violence relating to organized crime remains high in many regions in the world and, when linked to the illicit trafficking of arms and commodities, will derail efforts towards peace, human rights protection and sustainable development.
\end{quotation}
\end{CJK*}

This example expresses the UN's deep concern towards relentless violence, organized crimes, and homicides, which can cause devastating impact for international peace and security. The Chinese government has constantly sided with the UN in human rights protection as well as peacekeeping, and contributed its share in coordinating with the UN to combat violence-related crimes, human trafficking, and illegal smuggling of weapons. Therefore, when translating the ST into Chinese, the translator intentionally omits the hedge ``can'' to transform the possibility into a certainty, which serves to amplify the severe consequences brought by these crimes, and signals the Chinese government's shared stance with the UN.

\subsubsection{Modification}
Modification often occurs when the translator intends to raise or lower the definiteness of the statement, with the latter more often the case in RWG1 and RWG2. As suggested by example 3, despite the availability of the closest equivalence, the translators may opt for other hedging devices in the TT.

\begin{CJK*}{UTF8}{gkai}
\begin{quotation}
\noindent Example 3 (2018 report on the work of the government)

ST: 今年中央财政投入增加 300 亿元以上，比上年增长 20\%以上。

Gloss: This year, the central government will appropriate more than 30 billion yuan for this purpose, at least 20\% more than last year.

TT: This year, the central government will appropriate around 30 billion yuan for this purpose, at least 20\% more than last year.
\end{quotation}
\end{CJK*}

As illustrated in example 3, the central government plans to invest a large sum of fund into agriculture and rural regions, as part of its efforts to realize the ``revitalization of rural areas''. In the ST, ``\begin{CJK*}{UTF8}{gkai}300 亿元以上\end{CJK*}'' ( more than 30 billion yuan ) demonstrates the great importance attached to the development of rural areas by the government. However, this amount is only budgeted but not fixed, thereby leaving the possibility that the actual investment may be slightly lower than 30 billion. For the purpose of precaution, ``\begin{CJK*}{UTF8}{gkai}以上\end{CJK*}'' is not translated into its nearest equivalence ``more than'', but modified to be ``around'' in the TT.

%% file: discussion.tex
\section{Discussion}
To summarize, the corpus-based study of the translation of hedges in political documents reveals that hedges are more frequently used in English than in Chinese across our three sub-corpora. To be specific, for C-E translation, more hedging devices are found in the TTs than in the STs, while for E-C translation, more hedges appear in the STs than in the TTs. This finding points to a systemic difference regarding the general frequency of hedges in political texts between Chinese and English. Results in our study can find echoes in the previous studies, which suggested that English is ``over-hedged''~\citep{2022Fu} compared with many other languages, including Spanish, Iranian, Arabian, and Chinese across various types of discourse and texts~\citep{2012Jalilifar,2022Aljawadi,2019Alonso,2011Hu,2021Gong,2013Yang}.

\subsection{Linking Power Distance and Hedges}
This disparity can be attributed to different power distances between China and English-speaking countries. Power distance is an anthropological concept propose by Hofstede as part of his cultural dimensions theory~\citep{2011Hofstede}, which reflects the unequal power distribution in a community or society, and can be used to understand the interpersonal relationship between individuals with varying degrees of power. To measure power distance in a quantitative way, Hofstede proposed Power Distance Index (PDI). According to his research results, China is a country with high PDI, meaning that the decisions, opinions, and actions of the leading cadre are not likely to be challenged, and the hierarchical distinction between the leadership and ordinary citizens should be accepted. In contrast, English-speaking countries like the United Kingdom and the United States are gauged to have low PDI scores, meaning that members in these societies are ready to challenge hierarchy and voice dissenting opinions against the authority.

In the hierarchical Chinese society, the central government represents the highest authority, and political documents issued by the government institutions are thought to be assertive, authoritative, and unquestionable~\citep{2016Tang}. As hedges are mostly ``informal, less specific markers of probability and uncertainty''~\citep[p.240]{1988Biber} to make the propositional content ``fuzzier or less fuzzy''~\citep[p.195]{1973Lakoff}, it is not difficult to understand why hedges are not frequently adopted in the reports on the work of the government, which often feature formality and seriousness. On the contrary, the United Nations does not position itself as an authoritative institution, but an organization that encourages cooperation and solidarity with the mission to preserve world peace~\citep{2021Thorvaldsdottir}. Therefore, the annual reports of the UN are written in a more engaging manner, which may explain the relatively higher occurrence of hedging expressions.

\subsection{Influence of the Shifting Translation Norms and Policy Orientation}
Ever since the sociological turn in translation studies, norm has gained its foothold in describing and explaining translation phenomenon related to power, ideology, and culture~\citep[p.26]{1999Hermans}. 

Although no consensus has been fully reached on whether to define norm as a descriptive or prescriptive concept, and its relationship with psycholinguistic as well as cognitive aspects~\citep{2020Kotze}, norms can be regarded as shared values or rules that ``govern, identify and individualize the social order of the cultural system where translation takes place''~\citep[p.89]{2016Enriquez-Aranda}. As pointed out by \citet{2013Wunderlich,2016Yu,2020Xu}, norms are, far from being constant and stable, susceptible to change and evolve. They ``emerge, diffuse, become internalized, and, once established, become subject to change resulting in their strengthening, weakening, or even erosion''~\citep[p.20]{2013Wunderlich}.

This dynamic nature of norms is particularly suitable to account for the diachronic shift regarding the frequency of hedges in C-E translation. To illustrate, accuracy and faithfulness to the STs have long been the dominant norm of translation accepted by institutional translators working for the government, since the original political documents, which represent the official voice and stance, are attached with absolute authority~\citep{1983Cheng,2008Wang,2021Jia}. Under this context, translations of political documents have been required to resemble the STs as much as possible, even at the expense of understandability and acceptability~\citep{2021Jia}. In the extreme cases, even the word order should not be easily changed, and any detail should not be ignored~\citep{1983Cheng}.

However, this norm has been questioned by a group of senior translators. In the meanwhile, since ``going global'' was designated as an important national strategy at the 16th National Congress of the CPC, institutional translators are entrusted with the responsibility to balance faithfulness and reception, in order to better represent the Chinese voices, enhance China's soft power, and engage with the international society~\citep{2015Li,2021Pan2}. Much attention has been shifted away from a blind pursuit of ``faithfulness'' to the integration of accuracy and acceptability~\citep{2021Jia,2014Tong}. The translation of Chinese political texts is expected to approach the international political discourse, and make adjustments from the aspects of content, style, and manner~\citep{2021Jia}.

As indicated by our data analysis in the previous parts, we can identify a changing pattern concerning the frequencies of hedges in the translated political texts from Chinese to English. Their significant increase in the TTs during the past two decades is a vivid reflection of the shifting norms of the political text translation in recent years, giving support to previous findings that the translation of the Chinese political documents has become more target-oriented, reader-friendly, and easier to receive by the international community~\citep{2021Pan2,2004Zhang}.

\subsection{Gatekeeping and the Directionality of Translating Hedges}
What is out of our expectation is the directionality as an important factor that influences the general frequency of hedges: in the two C-E sub-corpora, we see a bigger difference between the STs and TTs in respect to the occurrence of hedges than the E-C sub-corpus. While almost all the hedging expressions are retained when being translated to Chinese, the rate of retention is much lower for the other direction (see table 3 and table 8), which means that the translator's intervention is more prominent in C-E translation compared with E-C translation in our corpora. The mediation efforts by the institutional translators are referred to as ``gatekeeping''~\citep{2012Pollabauer,1988Fujii,1998Wadensjo,2019Gu}, which has been used as a metaphor in translation studies to indicate the active role of translators guided by certain ideologies to intervene in the process of text reproduction. In the context of C-E translation of political documents, since the reports on the work of the government often contain major national policies, reforms, and strategies, they are of vital importance for both the domestic and international audience. Therefore, when they are translated into English, the translators are particularly careful in the delivery of the message. Translators as gatekeepers are expected to filter the messages concerning national image and interests in the STs by means of various handlings (i.e. addition, omission, modification), with an aim to both maintain political correctness and ensure that the translated texts are clearly understood and well-received by international readers. In contrast, the major concerns and objectives of the UN are international regional affairs. Issues related to China are seldom mentioned in the UN annual reports (in fact, China and the Chinese government are never mentioned in the 2018~2022 UN annual reports). Therefore, the gatekeeping function of translators is less obvious for E-C translation, manifested by the relatively high retention rate of hedges, and extremely low adoption of other translation methods
other than several cases of omission (see table 8).

%% file: conclusion.tex
\section{Conclusion}
Based on three customized corpora consisted of political documents from the reports on the work of the government and the UN annual reports, this study adopts a comparative approach to investigate how hedges are used in the STs and translated to the TTs. Overall, our analysis shows that hedges tend to appear more frequently in English political texts, be it translated English or original English, which points to a systemic difference regarding the frequency of hedges between the two distinct languages. In addition, directionality seems to play an important role in influencing what proportion of hedges in the STs will be translated into their nearest equivalents in the TTs. Our data reveal that the retention rate of hedges in E-C translation is higher than that in C-E translation of political documents, at least in the case of our three sub-corpora, which can be partly explained by the different degrees of gatekeeping efforts by the institutional translators.

Another major finding is that there is a noticeable diachronic increase in the occurrence of hedging devices in the English translations of the reports on the work of the government. For one thing, this change can be partly attributed to the increase of hedges in the STs; for another, there is still a portion of hedges that are intentionally added, and thus can be seen as the result of mediation by the translators. Such change occurs in parallel to the shifting policy that guides the outward translation of Chinese political texts, which turns from being ``source-oriented'' to ``target-oriented'' in recent years. Differences also exist in terms of the preferred category of hedges in the two C-E sub-corpora: while adaptors appear significantly more frequently in the reports roughly two decades ago, attribution shields, especially ``-based'' and ``based on'' serve as the most preferred hedging devices in the current years. Furthermore, our study finds that among the four identified handlings of hedges, retention is the dominant one, while omission is the least likely to be seen. Addition and modification are also actively deployed in C-E translation, but not used at all in E-C translation.

Although much attention has been paid to hedges since \citet{1973Lakoff}, studies devoted to the use of hedges in translated political texts remain scarce. This study, with the help of authentic and fit-for-purpose textual data, along with corpora-building tools, offers some insight into how hedges are used differently between the STs and the TTs, between C-E translation and E-C translation, as well as between two time periods. Additionally, the study attempts to explain these differences from a socio-cultural perspective, and reveals the integral relationship among translation, ideology, power, and culture. Nevertheless, restricted by the size of the corpora, our findings remain preliminary, and thus more future studies are awaited to offer systematic and in-depth discussion of hedging expressions in the translation of political documents.